\documentclass[11pt, a4paper]{google}

\usepackage{natbib}
\usepackage{hyperref} 
\usepackage{url}
\usepackage{authblk}  

\usepackage{graphicx}%
\usepackage{multirow}%
\usepackage{amsmath,amssymb,amsfonts}%
\usepackage{amsthm}%
\usepackage{mathrsfs}%
\usepackage[title]{appendix}%
\usepackage{textcomp}%
\usepackage{manyfoot}%
\usepackage{booktabs}%
\usepackage{algorithm}%
\usepackage{algorithmicx}%
\usepackage{algpseudocode}%
\usepackage{listings}%
\usepackage{float}
\usepackage[table]{xcolor}
\usepackage{adjustbox}
\usepackage{xfrac}
\usepackage{subcaption}

\definecolor{lightblue}{RGB}{212,232,248}
\definecolor{lightred}{RGB}{250,218,214}
\definecolor{lightgreen}{RGB}{235,243,231}
\definecolor{orange}{RGB}{255,127,0}

\newcommand{\fnm}[1]{#1}
\newcommand{\sur}[1]{#1}
\newcommand{\orgname}[1]{#1}
\newcommand{\orgaddress}[1]{#1}
\newcommand{\city}[1]{#1}
\newcommand{\country}[1]{#1}

\newif\ifArxiv
\Arxivtrue

\begin{document}

\title{Tree crop mapping of South America reveals links to deforestation and conservation}

\author[1,5]{\fnm{Yuchang} \sur{Jiang}}
\author[1]{\fnm{Anton} \sur{Raichuk}}
\author[2]{\fnm{Xiaoye} \sur{Tong}}
\author[5]{\fnm{Vivien Sainte Fare} \sur{Garnot}}
\author[2]{\fnm{Daniel} \sur{Ortiz-Gonzalo}}
\author[3]{\fnm{Dan} \sur{Morris}}
\author[4]{\fnm{Konrad} \sur{Schindler}}
\author[5]{\fnm{Jan Dirk} \sur{Wegner}}
\author[1]{\fnm{Maxim} \sur{Neumann}}

\affil[1]{\orgname{Google DeepMind}, 
\orgaddress{\city{Z\"urich}, \country{Switzerland}}}
\affil[2]{\orgname{University of Copenhagen}, 
\orgaddress{\city{Copenhagen}, \country{Denmark}}}
\affil[3]{\orgname{Google Research}, 
\orgaddress{\city{Seattle, WA}, \country{USA}}}
\affil[4]{\orgname{ETH Z\"urich}, 
\orgaddress{\city{Z\"urich}, \country{Switzerland}}}
\affil[5]{\orgname{EcoVision Lab, DM3L, University of Zurich}, 
\orgaddress{\city{Z\"urich}, \country{Switzerland}}}

\correspondingauthor{yuchang.jiang@uzh.ch, maximneumann@google.com}

\begin{abstract}
Monitoring tree crop expansion is vital for zero-deforestation policies like the European Union's Regulation on Deforestation-free Products (EUDR). However, these efforts are hindered by a lack of high-resolution data distinguishing diverse agricultural systems from forests. Here, we present the first 10 m-resolution tree crop map for South America, generated using a multi-modal, spatio-temporal deep learning model trained on Sentinel-1 and Sentinel-2 satellite imagery time series. The map identifies approximately 11 million hectares of tree crops, 23\% of which is linked to 2000-2020 forest cover loss.
Critically, our analysis reveals that existing regulatory maps supporting the EUDR often classify established agriculture, particularly smallholder agroforestry, as ``forest''. 
This discrepancy risks false deforestation alerts and unfair penalties for small-scale farmers. 
Our work mitigates this risk by providing a high-resolution baseline, supporting conservation policies that are effective, inclusive, and equitable.
\end{abstract}

\maketitle

\section*{Introduction}
Tree crops, encompassing perennial woody plants from temperate species 
to tropical commodities, 
are at the center of a global debate. While contributing significantly to rural economies and global trade, their expansion in tropical regions is a high-risk driver of deforestation, forest degradation, and associated biodiversity and carbon losses \citep{vijay2016palmDeforestation,meijaard2020environmentalImpactPalm,campera2022abundanceCoffee}. 
This expansion has encroached on protected areas and indigenous territories \citep{kalischek2022_cocoaETH}, prompting significant policy responses to govern sustainable supply chains. A key example is the European Union’s Regulation on Deforestation-free Products (EUDR), adopted in 2023, which requires importers of such commodities into the EU market to demonstrate that their products are not sourced from former forest lands converted after December 31, 2020. Implementating such policies underscores the need for reliable high-resolution datasets that distinguish natural forests from agricultural tree systems, to ensure fair, transparent, and enforceable sustainability standards \citep{berger2025earth}.

A major gap in land monitoring efforts is the absence of tree crops as a distinct land-cover category \citep{van2025beyondImperfectMap}.
Foundational works like \cite{hansen2013high_Hansen_forestloss} and \cite{shimada2014ALOS-forestMap} focus on differentiating forests from areas not covered by trees, without identifying tree crops within them. More recent efforts \citep{vancutsem2021long_TMF, lesiv2022global_lesiv2015}, including the JRC Global Map of Forest Cover (GFC) for 2020 \citep{bourgoin2024JRCglobalForestMap, bourgoin2024JRCglobalForestTypes}, provide crucial baselines for forest monitoring. However, these datasets are generally optimized for forest extent rather than detailed land-use classification, and may not consistently differentiate tree crops from other forest types, particularly in complex agroforestry systems \citep{verhegghen2024EUDRCaseStudy, van2025beyondImperfectMap}. Such ambiguities can generate false deforestation alerts and may have unintended consequences for local livelihoods, particularly for smallholder farmers.
Addressing this gap is essential for improving regulatory compliance and advancing the United Nations Sustainable Development Goals (SDGs)---notably SDG 15 (Life on Land), SDG 12 (Responsible Consumption and Production), and SDG 1 (No Poverty)---by promoting sustainable land use and equitable treatment of smallholders.
While numerous studies have mapped single tree crop species \citep{descals2024global_palmDescals,danylo2021palmSEA, lin2021orchard,adrah2025integrating, peng2024tea, wang2023rubber, sheil2024rubber, descals2023high_coconutDescals, kalischek2022_cocoaETH, maskell2021coffeeVietnam}, these efforts are typically regional and use diverse methodologies, preventing the harmonized, continental-scale assessment required for comprehensive monitoring.

Mapping tree crops at a continental scale is challenging. First, it is difficult to distinguish tree crops from land cover patterns with similar spectral and textural signatures \citep{yang2025mapping}. Tree crops (e.g., oil palms) and timber plantations (e.g., eucalyptus) are often planted in evenly spaced rows. In contrast, perennial crops like cocoa and coffee in smallholder systems are more heterogeneous, with irregular canopy structure \citep{batista2022optical}. Furthermore, many smallholder plantations are highly fragmented, occupying fractions of a hectare. Monitoring them requires satellite imagery with sufficient spatial, spectral and temporal resolution, and classification methods that account for the complexities of such data.
%
%
Another key challenge is the fragmented and uneven distribution of reference data needed to train accurate models. The current data landscape comprises various efforts at different scales, each with limitations. Regional and continental initiatives, such as MapBiomas \citep{souza2020reconstructingMapBiomasBrazil}, provide valuable annual land-use monitoring, but tree crops are not consistently represented. The global Spatial Database of Planted Trees (SDPT \citep{harris2019SDPTv1,richter2024SDPTv2}) offers crucial location data but is incomplete and varies in quality. While regional datasets exist for specific crop types \citep{descals2024global_palmDescals,danylo2021palmSEA,lin2021orchard,adrah2025integrating,peng2024tea,wang2023rubber,sheil2024rubber,descals2023high_coconutDescals,kalischek2022_cocoaETH,maskell2021coffeeVietnam}, these efforts are fragmented across countries, differ in methodology, and vary in spatial resolution and accuracy. 
To address this, we combined accessible sources, including regional tree crop maps and global land cover datasets into a unified reference dataset that we used for model training and mapping.

Here, we analyze the expansion of tree crops across South America and its links to historical forest cover loss and conservation. 
We generated the first 10 m-resolution, continentally consistent tree crop map by training a model on the integrated diverse reference datasets and leveraging multi-temporal satellite imagery. Building on advances in data-driven modeling for vegetation monitoring \citep{brandt2025high,lin2021orchard,kalischek2022_cocoaETH,descals2023high_coconutDescals,descals2024global_palmDescals,garnot2021UTAE,tarasiou2023TSViT,pazos2024planted}, we trained a deep learning model using Sentinel-1 (synthetic aperture radar) and Sentinel-2 (optical) time series to distinguish tree crops from other land uses.
We quantify the extent of tree crops associated with historical forest cover loss and examine their spatial relationships with protected areas. 
Finally, we compare our results with existing forest cover products used in policy contexts such as EUDR. By capturing smallholder-dominated tree crop systems often indistinguishable in broader forest datasets, our map complements existing data to provide stronger, more transparent evidence for sustainability monitoring, helping reduce compliance burdens on smallholders and support equitable, evidence-based conservation policies.

\section*{Results}

\subsection*{A high-resolution tree crop map of South America}
\begin{figure}
   \centering
        \includegraphics[width=\linewidth]{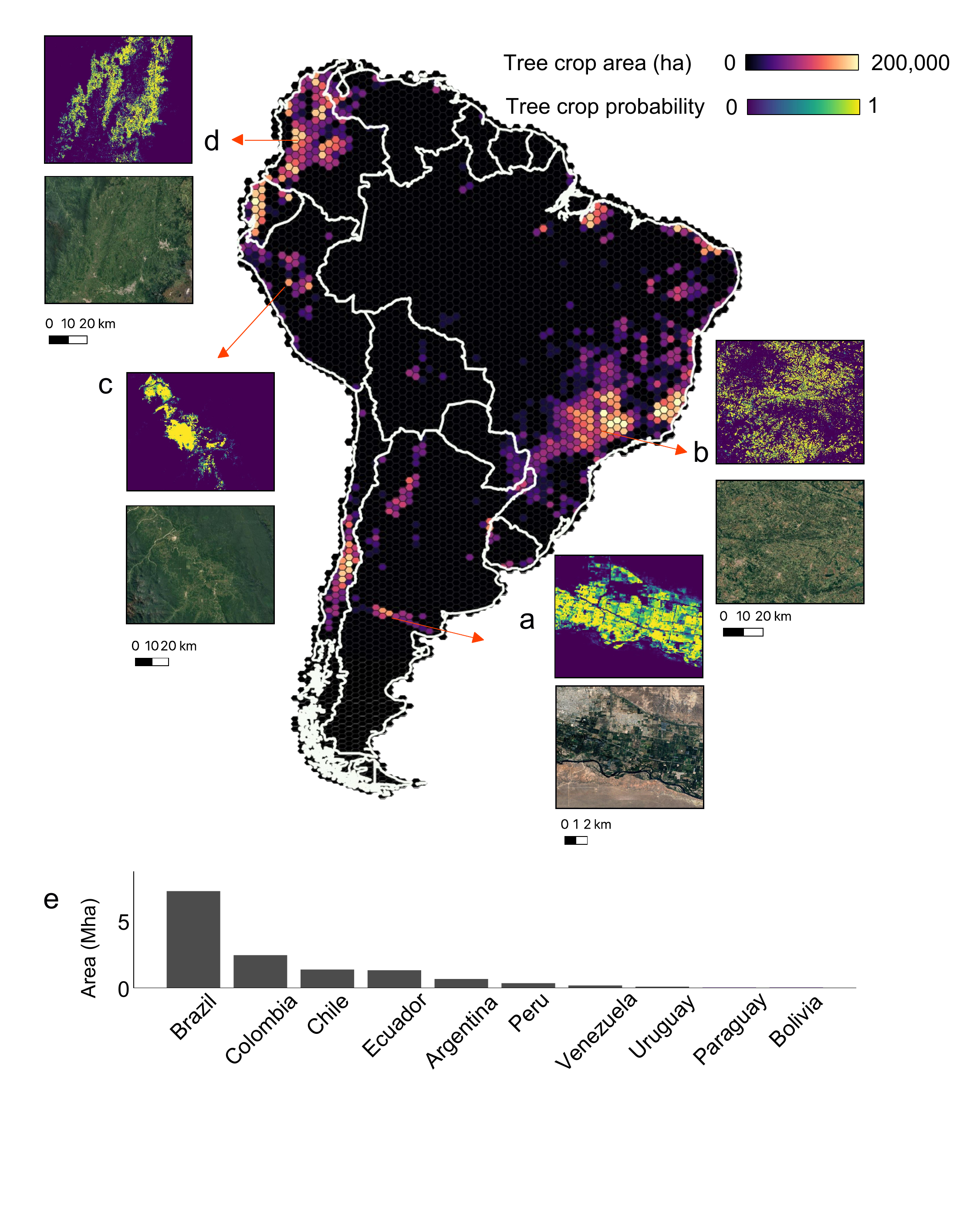}
        \caption{Tree crop map for South America. The main map displays a hexagon-based overview (80km per hexagon side), where brighter colors indicate higher tree crop density. Detailed maps a--d show zoomed-in views of the tree crop probability in the tree crop regions of (a) Argentina (mainly apple orchards), (b) Brazil (large-scale coffee plantations), (c) Peru (palm plantations), and (d) Colombia (smallholder coffee farms). (e) Tree crop area per country.}
        \label{fig1:final_map}
\end{figure}

We used the dense satellite imagery time series from both Sentinel-2 and Sentinel-1 to generate seasonal composites covering all continental South America. We trained a multi-modal spatio-temporal vision transformer (MTSVIT) model on a unified reference dataset, harmonized from multiple sources, to generate a 10$\,$m-resolution tree crop map for 2020 (Figure \ref{fig1:final_map}a). On a held-out test set, the model achieved an F1 score of 89.5\% on tree crops, with 87.0\% recall and 92.1\% precision.

The high-resolution map reveals distinct spatial patterns and geographic clustering of tree crops. Total tree crop areas by country are shown in Figure \ref{fig1:final_map}b. Brazil dominates with 6.64 million ha, concentrated in Minas Gerais, São Paulo, and Paraná. In Colombia, tree crops occupy 1.39 million ha and form a well-defined cluster in the Coffee Triangle, capturing the fine-grained mosaic of smallholder coffee farms. The map highlights additional hotspots: oil palm and cocoa plantations in Loreto and Ucayali (Peru); tree crops concentrations in the coastal provinces in Ecuador; and temperate fruit production clusters in the Maule Valley (Chile) and the Villa Regina region of northern Patagonia (Argentina). Discernible patches are also present in Bolivia’s lowlands, especially near Santa Cruz.
These results provide the first detailed, continentally-consistent, high-resolution overview of tree crop distributions across South America.

We performed a rigorous accuracy assessment using established probability-based sampling protocols \citep{olofsson2014good, olofsson2020mitigatingOmissionError} to generate unbiased area estimates. 
Tree crops are a rare class ($<1$\% of the continent), and this imbalance affects the stability of omission errors and area-adjusted accuracy estimates. We followed recommendations \citep{olofsson2020mitigatingOmissionError} to perform area-based accuracy adjustment only for countries where tree crops occupy more than 0.5\% of the national area. 
The evaluation on an independent validation dataset yielded an area-adjusted user’s accuracy of 83\% and a producer’s accuracy of 66\% for tree crops, with an overall accuracy of 99.6\%. Country-specific metrics are similar (Extended Data Table \ref{tab:area_estimate_country}). We used the adjusted areas from these countries to compute a scaling factor ($\frac{\text{adjusted area}}{\text{initial area}} = 1.26$), which was applied to adjust areas in countries with stronger class imbalance. After adjustment, the total tree crop area across South America is estimated to be 10.99 million hectares.

\subsection*{Nearly \sfrac{1}{4} of all tree crops grow in forest cover loss area}

\begin{figure}[H]
   \centering
        \includegraphics[width=\linewidth]{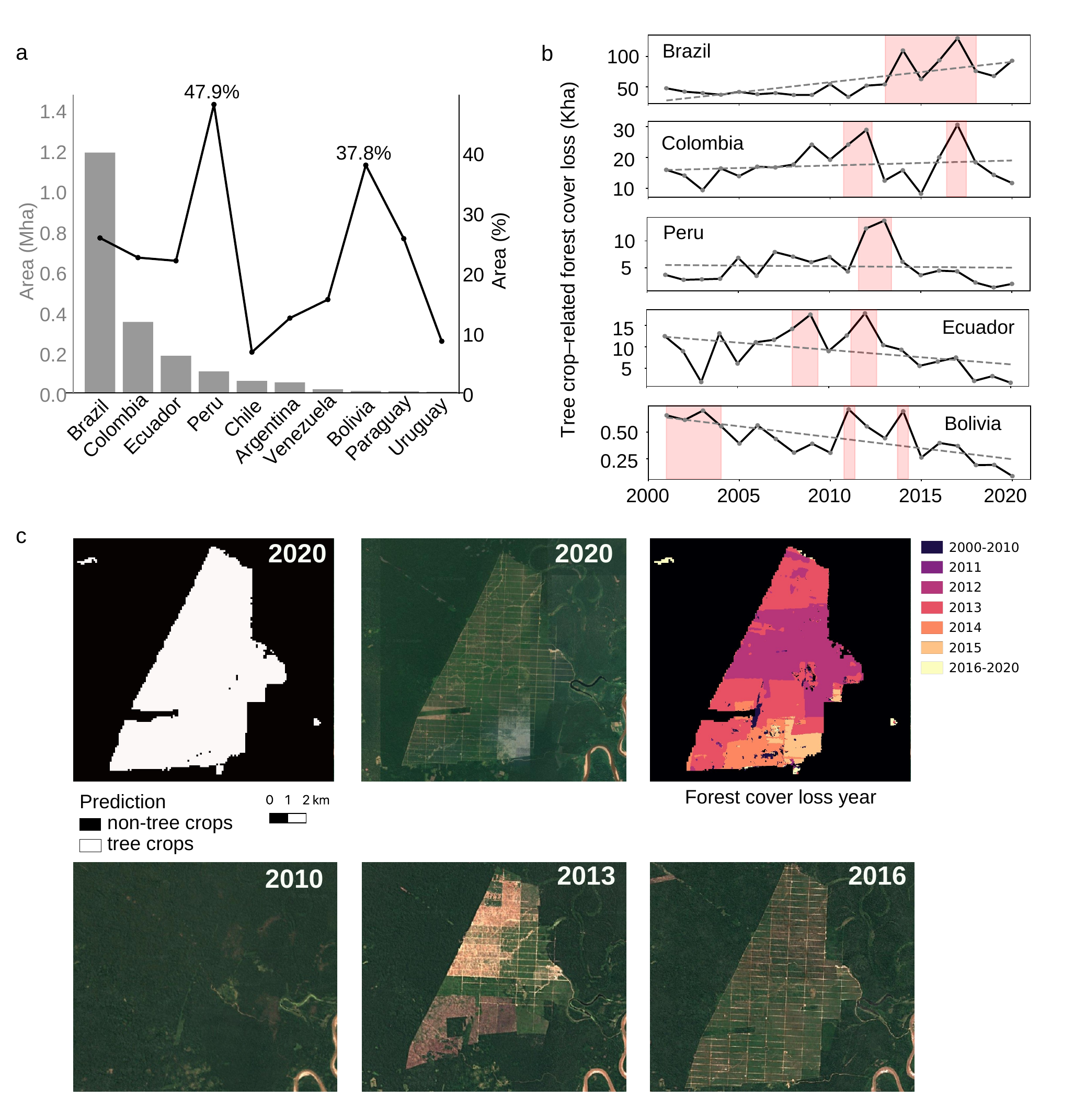}
        \caption{Tree crops in the forest cover loss areas.
        a) Tree crop area within forest cover loss zones (left axis, ha) and its proportion relative to total tree crop area (right axis, \%) by country. While higher tree crop area generally corresponds to greater overlap with forest cover loss, Peru stands out as an outlier with a high relative proportion despite a smaller absolute area.
        b) Temporal patterns of tree crops within forest cover loss in South America. The line chart shows the annual tree crop area inside forest cover loss per country.
        c) Example of large-scale oil palm expansion in Ucayali, Peru.
        }
        \label{fig2:forest_cover_loss}
\end{figure}

There is often a strong association between tree crop expansion and forest cover loss. 
We compared our tree crop map with the global forest cover loss (GFCL) dataset for 2000–2020 \citep{hansen2013high_Hansen_forestloss}.
Across all studied countries, approximately 23\% of the mapped tree crop area overlaps with forest cover losses during this period. In absolute terms, Brazil shows the largest extent of tree crops on land that experienced recent forest cover loss (1.19 million ha), followed by Colombia (0.35 million ha). Countries with larger tree crop extents generally exhibit greater overlap with forest cover loss (Figure \ref{fig2:forest_cover_loss}a), and the spatial clustering resembles the overall distribution of tree crops (Supplementary Figure \ref{supplementary_fig:forest_loss_spatial}).
In relative terms, Peru (47.9\%) and Bolivia (37.8\%) show the highest proportions of tree crops established within areas of historical tree cover removal.

To understand the timing of this conversion, we intersected our 2020 tree crop map with annual forest cover loss layers from 2000 to 2020  \citep{hansen2013high_Hansen_forestloss}. This analysis captures country-level temporal patterns of forest cover removal within current tree crop regions (Figure~\ref{fig2:forest_cover_loss}b). 
The resulting time series reveals distinct national patterns: Brazil shows a gradual, persistent increase in tree crop–related forest cover loss over two decades, whereas Ecuador and Bolivia exhibit lower magnitudes and decreasing trends with episodic peaks. Beyond these differences, an overall increase is observed between 2008 and 2013 across the continent.
In Peru, a pronounced peak between 2010 and 2015 corresponds to large-scale oil palm development in Ucayali and Loreto, aligning with previous reports \cite{furumo2017characterizingOilPalm}. Spatial analysis of forest loss years (Figure~\ref{fig2:forest_cover_loss}c) shows that most large oil palm plantations in Ucayali were established during 2013 to 2016, a trend confirmed by historical satellite imagery and consistent with previous findings \citep{glinskis2019quantifyingPeruPalm2016}. 
In Peru, a pronounced peak between 2010 and 2015 corresponds to large-scale oil palm development in Ucayali and Loreto; spatial analysis of forest loss years (Figure~\ref{fig2:forest_cover_loss}c) shows that most large plantations in Ucayali were established during this period, a trend confirmed by historical satellite imagery and consistent with previous studies \citep{glinskis2019quantifyingPeruPalm2016, furumo2017characterizingOilPalm}.
In Colombia, forest cover loss rose after the 2016 peace process \citep{clerici2020ColombiaConflictdeforestation}.

This comparative analysis has limitations. The association with forest cover loss may be underestimated where plantations predate the year 2000, or overstated when tree crops are abandoned in favor of alternative non-forest land uses. As GFCL \citep{hansen2013high_Hansen_forestloss} is based on canopy cover thresholds, we further sampled a random subset of points from the overlap between our tree crop map and the GFCL layer and visually examined them using historical high-resolution imagery. This assessment revealed that 36\% cases correspond to forest conversion to tree crops (e.g., large-scale oil palm), while 35\% of samples reflect replanting or management within existing, typically smallholder plantations (see Supplementary Section~\ref{sanity_check_tc_forest_cover_loss}). Given that tree crops like oil palm typically persist for at least 25 years before replanting \citep{furumo2017characterizingOilPalm}, most canopy removals associated with these large-scale systems likely represent new establishment rather than rotation.

\subsection*{Tree crop cultivation concentrates at protected area boundaries}

Protected Areas (PAs) play a key role in conserving biodiversity and mitigating deforestation, though their effectiveness varies across management categories, from strictly protected reserves to multiple-use landscapes where limited human activities are permitted.
We assessed the spatial relationship between tree crops and PAs to evaluate potential encroachment. 
Brazil shows the largest overlap, with 150,190 ha of tree crops located within PAs (3.3\% of its total tree crop area), followed by Colombia (29,240 ha; 1.9\%), Argentina (3,286 ha; 0.8\%), and Ecuador (402 ha; 0.05\%).
As PA boundaries are vulnerable to human pressure, analyzing tree crop distribution along them helps to understand how tree crop expansion interacts with conservation areas. 

\begin{figure}[!ht]
   \centering
        \includegraphics[width=\linewidth]{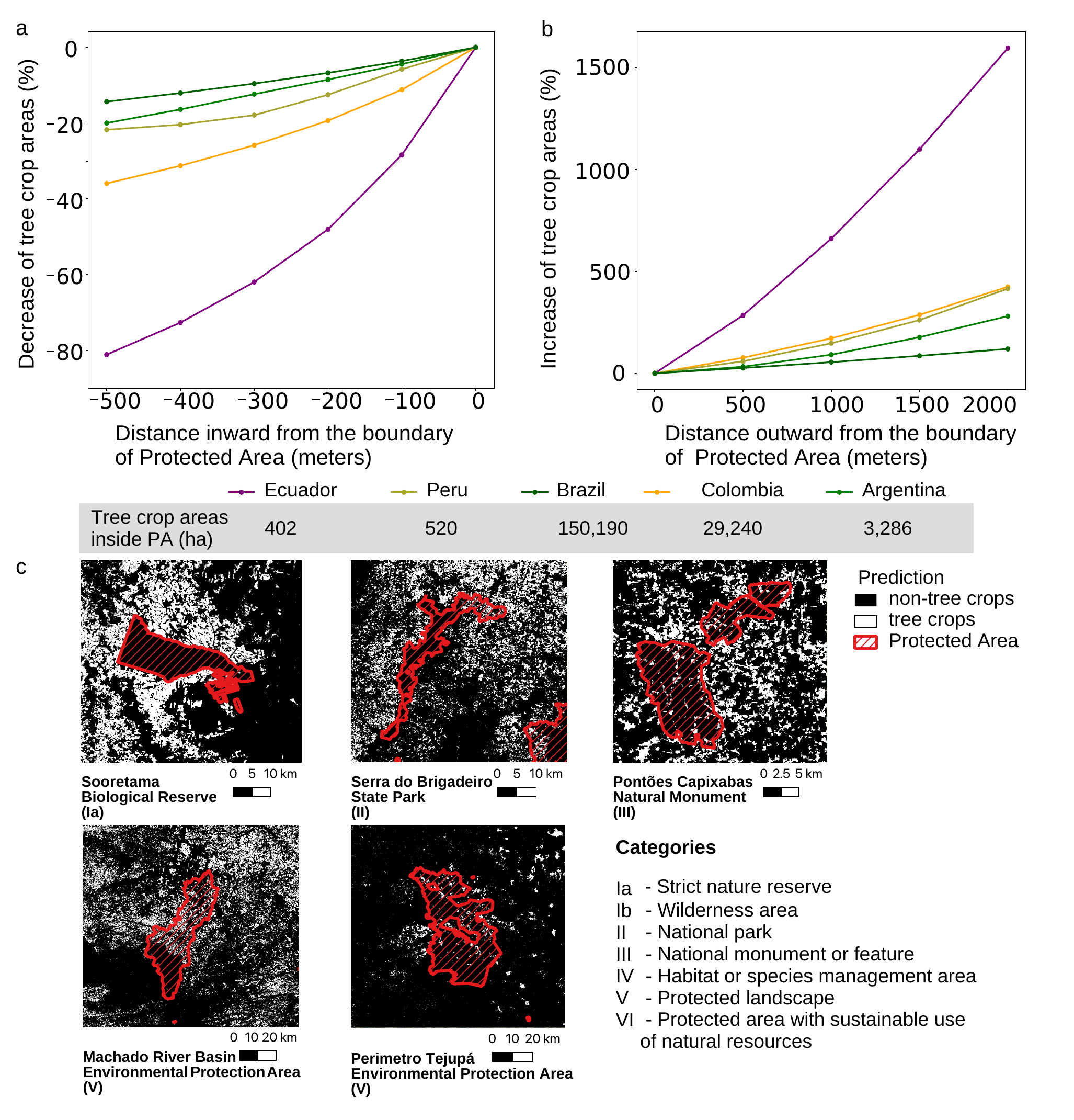}
        \caption{Tree crops in the Protected Areas.
        a) Decrease in tree crop area with increasing distance from the boundary into Protected Areas.
        b) Increase in tree crop area with increasing distance outward from the boundary of the Protected Areas.
        c) Examples of selected Protected Areas and PA category names. The red rectangles are Protected Areas, white indicates the predicted tree crop areas while black indicates non-tree crop areas.
        }
        \label{fig3:protected_areas}
\end{figure}

To investigate this distribution, we conducted a buffer analysis using negative (inward) and positive (outward) buffers. For each country, we quantified changes in tree crop density across the boundary by comparing their proportion within successive buffer distances to the one inside PAs (Figure~\ref{fig3:protected_areas}a,b).
The results reveal a clear boundary effect: in all five countries, tree crop presence declines sharply toward the interior of PAs. This is particularly pronounced in Ecuador, where tree crop area decreases by nearly 80\% within the first 500$\,$m inside the boundary. Similar but less steep declines are observed in Colombia, Peru, Argentina, and Brazil, indicating that tree crops inside PAs are predominantly concentrated near their peripheries rather than deep within core zones. Conversely, tree crop density increases rapidly outside the PA boundaries, reaching between 2$\times$ (Brazil) and 15$\times$ (Ecuador) the density inside the PA within the first 2$\,$km. 
These patterns demonstrate that tree crop cultivation is spatially clustered along PA edges.
We also find that most tree crops within PAs occur in lower protection categories, such as protected landscapes (\textit{Category V}), rather than in strictly protected reserves such as nature reserves (\textit{Category Ia}) (Figure~\ref{fig3:protected_areas}c). \textit{Category Ia} areas rarely have tree crops extending into the interior, whereas \textit{Category V} areas exhibit substantial presence; indicating that higher levels of protection are less affected by tree crop expansion.

\subsection*{Enhanced mapping of smallholder farms for EUDR}

\begin{figure}[!ht]
   \centering
        \includegraphics[width=\linewidth]{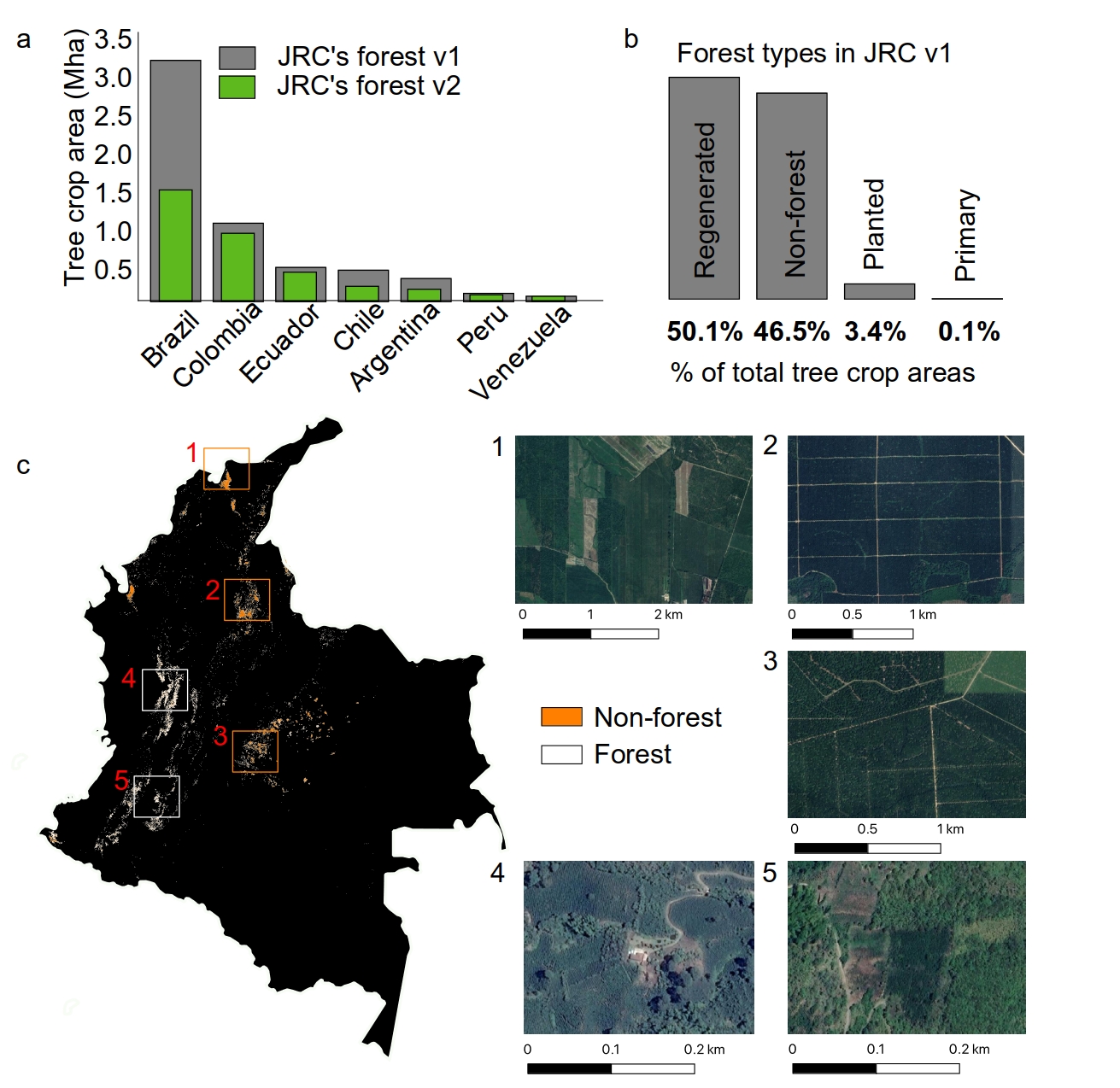}
        \caption{Addressing EUDR in comparison of JRC maps.
        a) Comparison of tree crop representation in version 1 and version 2 of the JRC Global Forest Cover 2020 map. 
        b) The bar chart shows the distribution of JRC’s Global Forest Type 2020 classes within our mapped tree crop areas.
        c) Zoomed-in examples from Colombia comparing our tree crop map with the JRC Global Forest Cover 2020 (v2). Orange indicates areas classified as non-forest by JRC, while white indicates those classified as forest. Large oil palm plantations are generally mapped as non-forest (windows 1–3 with latitude and longitude: [10.672, –74.240], [7.264, –73.889], [3.916, –73.385]), whereas smaller fields, such as coffee, are often classified as forest (windows 4–5 with latitude and longitude [5.321, –75.733], [2.499, –76.042]).
        }
        \label{fig:eudr_jrc_compare}
\end{figure}

We examined how current regulatory support maps, often referenced in EUDR-related assessments \citep{Meinevan2025european},  represent tree crops. While EUDR compliance requires rigorous due diligence based on geolocation data and cannot rely solely on remote sensing maps, high-quality maps provide essential contextual information for risk assessment. We compared our tree crop map with three JRC products: the Global Forest Cover 2020 map (GFC), versions 1 and 2 \citep{bourgoin2024JRCglobalForestMap}; and the Global Forest Type 2020 map (GFT) \citep{bourgoin2024JRCglobalForestTypes}. 
We observed notable differences in how these maps characterized areas identified as tree crops in our study. GFC v2 designates 3.27 Mha of our mapped tree crops as forests, a smaller area of disagreement compared to v1 (5.50 Mha), indicating improved distinction between forests and certain tree crop systems in v2. Similarly, the GFT product classifies 50.1\% of the areas we identified as tree crops as ``naturally regenerated forest'' (Figure \ref{fig:eudr_jrc_compare}b).
The divergence between forest extent maps and land-use specific maps highlights potential challenges for EUDR risk assessments, particularly if established agriculture is classified as forest.

These classification differences are unevenly distributed. In Brazil and Chile, the differences observed in v1 were substantially reduced in GFC v2 (53.4\% and 50.0\%, respectively) (Figure \ref{fig:eudr_jrc_compare}a). However, Colombia, the second-largest tree crop producer, experienced only a 12.4\% decrease in the area of divergence.
In Colombia, large oil palm plantations are generally classified as non-forest in GFC v2; whereas smaller plantings, like those in the Coffee Triangle, frequently remain labeled as forest (Figure \ref{fig:eudr_jrc_compare}c). Five examples illustrate this contrast: examples 1–3 show large-scale oil palm plantations where GFC v2 and our map agree on the non-forest classification, whereas examples 4-5 depict small-structured coffee plots interspersed with shade trees. Their mixed canopy structure is more challenging and is often classified as forest in global forest products. Comparison with reference samples used for our accuracy assessment shows that GFC v2 exhibits a misclassification rate of 10\% (reduced from 43\% in v1), while our map achieves 4\%. Most remaining inconsistencies occur in smallholder-dominated areas. Such discrepancies can lead to false deforestation signals if management activities are mistaken for forest clearing,  disproportionately affecting smallholder agriculture and ecologically desirable shade tree systems, underscoring the need for complementary, land-use specific data layers.

\section*{Discussion}

This study provides the first 10$\,$m-resolution tree crop map of South America, explicitly representing an important agricultural land-use class that previously was largely not accounted for in continental-scale products. By combining multi-temporal Sentinel-1 (cloud-penetrating radar) and Sentinel-2 (multi-spectral) satellite imagery with advanced deep learning, we identify major production regions and fine-grained tree crop systems across diverse ecosystems. 
The integration of multi-modal, multi-temporal inputs is well-suited for the heterogeneous and dynamic characteristics of tree crop systems, which traditional mapping approaches often cannot handle. This methodology provides a scalable framework for the challenging tree crop mapping task that can be extended to other globally significant regions.

The resulting map offers new insights into the association between tree crops, forest cover change, and conservation.
Our analysis reveals substantial spatial overlap between tree crop areas and regions of forest cover loss since 2000, based on satellite-derived forest loss products that estimate tree cover removal \citep{hansen2013high_Hansen_forestloss}. While these products provide the most comprehensive global record of tree cover dynamics, they measure biophysical tree cover loss, which may not always equate to the deforestation of natural forests. Acknowledging these limitations, the strong spatial overlap suggests that tree crop expansion is linked to forest cover loss, likely in combination with other interacting drivers such as infrastructure development, agricultural frontier expansion, and policy incentives \citep{armenteras2017deforestation, curtis2018classifyingDrivers}.
A pronounced regional increase of tree crops area occurred between 2010 and 2015, coinciding with a wave of biofuel initiatives across the continent. Biodiesel programs introduced in Brazil, Colombia, Peru and Ecuador during that period encouraged oil palm cultivation, likely contributing to the expansion \citep{furumo2017characterizingOilPalm}.
Beyond this trend, the map reveals distinct national dynamics. Brazil showed a gradual but persistent trajectory, whereas Peru and Bolivia experienced rapid conversion concentrated along emerging commodity frontiers. In Bolivia, the overlap of tree crops with forest cover loss reveals characteristic ``fishbone'' patterns along road infrastructure, particularly along the Santa Cruz–Cochabamba highway, a well-documented deforestation frontier \citep{steininger2001tropicalDeforestation, van2006drivesBolivia}. In Peru, our map confirms a period of industrial oil palm expansion between 2010 and 2016 in Loreto and Ucayali\citep{gutierrez2013PeruPalmannualmap2010, glinskis2019quantifyingPeruPalm2016}. In Colombia, increased forest cover loss after 2016 may be linked to the peace agreement that followed decades of armed conflict, which opened previously inaccessible areas to agricultural development \citep{clerici2020ColombiaConflictdeforestation}.
These findings demonstrate how the new map enables spatially and temporally explicit investigations of tree crop expansion.
Our analysis of protected areas complements this: while strictly protected reserves remain relatively intact, tree crops are concentrated along PA boundaries, particularly those in lower protection categories.
This suggests that these boundary regions are indeed under high pressure, potentially indicating deforestation ``spillover'' effects (leakage), where conservation efforts within PAs displace agricultural expansion to immediate surroundings \citep{fuller2019:pa-spillover}.
These observed patterns warrant further investigation to understand the precise drivers of this expansion and to rule out potential artifacts in the underlying land cover or forest loss datasets.

Arguably the most critical application of this map is to inform conservation policy, particularly the EUDR. Effective implementation requires a clear distinction between natural forests and agricultural production systems.
Existing forest cover datasets \citep{bourgoin2024JRCglobalForestMap, bourgoin2024JRCglobalForestTypes, neumann2025natural}, provide baselines for identifying forest around the 2020 EUDR cut-off date. Our map complements these datasets by adding essential land-use specificity---distinguishing perennial agriculture (plantations and agroforestry) from natural forests. Together, these products can offer stronger, auditable evidence for due diligence, supporting fairer treatment of smallholders, improving leakage detection, and enabling transparent EUDR implementation.
This distinction is vital because forests and agroforestry systems often share similar spectral signatures and both exceed canopy-cover thresholds, yet they represent different land uses \citep{Meinevan2025european}.
Our analysis highlights that areas identified as tree crops in our map are frequently mapped as ``naturally regenerating forest'' in broad forest cover products, especially in regions dominated by smallholder systems. Similar observations have been reported in Côte d’Ivoire and Indonesia, where tree crops were frequently classified as forest in existing datasets \citep{verhegghen2024EUDRCaseStudy,van2025beyondImperfectMap}. While these map-to-map comparisons must be interpreted cautiously as both maps contain uncertainties, they highlight a significant implementation risk: without land-use nuance, areas managed under long-standing, ecologically beneficial practices risk being flagged as deforestation if management activities (like replanting) are mistaken for forest clearing. 
This could place disproportionate burdens on smallholders who already face high compliance costs and limited capacity for geospatial verification \citep{Meinevan2025european, leimona2024sustainabilityMeinevan}.
Our map provides the high-resolution data necessary to distinguish these systems, strengthening the evidence base for equitable and effective EUDR compliance and advancing broader SDGs.

While the map provides a consistent, high-resolution inventory of tree crops throughout South America, it has limitations that should be addressed in future work.
First, although the map improves the capture of agroforestry systems over previous products, these systems remain challenging to map, not least due to limited field data. Expanding ground observations, especially for shaded and heterogeneous tree crop systems, is crucial for policy applications. Incorporating higher-resolution optical imagery and advanced radar processing could also improve detection.
Second, our analysis of historical links to forest cover change relies on external datasets \citep{hansen2013high_Hansen_forestloss}; uncertainties in these inputs propagate into our analysis, necessitating cautious interpretation of temporal dynamics.
Third, comparisons between our map and existing forest products (Figure \ref{fig:eudr_jrc_compare}) are based on map-to-map analysis rather than a comparison against ground truth specifically designed to evaluate these differences. Future work should include targeted validation efforts in areas of disagreement, particularly in complex smallholder landscapes, to better characterize the different mapping approaches (see Supplementary Materials for qualitative examples).
Fourth, the area adjustment approach \cite{olofsson2020mitigatingOmissionError} is most reliable for major producer countries. In countries with very small fractions of tree crop land, extreme imbalance introduces uncertainty, requiring improved area-adjustment strategies.
Fifth, a continent-wide map with a unified methodology provides consistency but may disregard regional peculiarities. Variations in landscape structure and management practices may lead to local over- or underestimation. For specific regions, calibrating the map to that context is advisable. For example, the optimal thresholds for converting tree crop probabilities into binary presence/absence maps are likely to vary across regions and use cases and can be re-calibrated for best results. We will release the map of tree crop probabilities to support local refinement and adaptation.

\section*{Methods}

\subsection*{Input data}

We used multispectral imagery from Sentinel-2 and dual-polarization Synthetic Aperture Radar (SAR) data from Sentinel-1 from ESA's Copernicus program as the model inputs. We used Level-2A products of Sentinel-2 from the year 2020 \citep{esa2015:s2-user-handbook}. These images were filtered using a cloud mask, discarding images with cloud cover exceeding 40\%. Valid pixels were aggregated into seasonal images by taking the median value of the time series over consecutive 3-month periods (only 0.005\% of locations remained permanently masked out due to cloud cover). We used all 10 spectral channels with 10m or 20m GSD (which are sensitive to the vegetation), with the 20m-channels resampled to 10m using Nearest Neighbor method.
For Sentinel-1 \citep{esa2022:s1-algo}, we used both the VV and VH polarizations, from both ascending (VV/VH${}_a$) and descending (VV/VH${}_d$) Sentinel-1 orbit directions. We concatenated the channels from ascending and descending passes and included the local incidence angle as additional channel.
Seasonal aggregation was performed similar to Sentinel-2, except that no cloud filtering was applied, since C-band SAR was minimally affected by clouds. Instead of the median we took the mean of the 3-month time series to reduce the speckle effect common in SAR data.

We compiled a comprehensive reference dataset by harmonizing multiple public data sources. The dataset has eight land cover classes, including tree crops, natural forest, planted forest, other vegetation, as well as built areas, water, ice/snow, and bare ground. These classes are intended to be comprehensive, i.e., every surface pixel can be assigned to one of these classes. 
A summary of data sources used is provided in supplementary Table \ref{tab:tree-crop-sources} and \ref{tab:other-sources}.
All reference data were resampled to 10m spatial resolution to match the spatial resolution of the input satellite data.

\subsection*{Training data sampling}
To construct the training dataset, we employed a sampling strategy designed to ensure adequate representation of all classes while maintaining geographic diversity. We performed random stratified sampling across South America, stratifying by available land-use classes within each 100$\times$100 km${}^2$ area. Crucially, we augmented these locations with the sample locations of known tree crops derived from our reference data compilation (supplementary Table \ref{tab:tree-crop-sources}) to imrpove the models' ability to distinguish these classes. The sampled locations served as center points to extract reference data and input satellite patches with dimensions 1280 meters $\times$ 1280 meters (128 $\times$ 128 pixels). This resulted in a dataset with 133,697 samples, with a total coverage of $219,049$ km$^2$, or about $1.2\%$ of the continent's area. Each sample includes the seasonal image patches of Sentinel-2 and Sentinel-1 (4 image patches for Sentinel-2 and 4 image patches for Sentinel-1). For the evaluation of model training performance and hyperparameter tuning, we performed geographic splitting of the samples by first dividing the study area into 100$\times$100 km\textsuperscript{2} cells and then randomly assigning these cells to the \textit{training}, \textit{validation}, and \textit{test} sets using a 8:1:1 ratio.

\subsection*{Deep learning framework}
Using the constructed dataset, we adapted and applied the Multi-modal Temporal Spatial Vision Transformer (MTSViT) to extract features from spatio-temporal satellite inputs and generate tree crop predictions (see Supplementary Figure \ref{model_overview} with training details). Sentinel-1 and Sentinel-2 image time series were first partitioned into 3D tokens ($1 \times 8 \times 8$: one temporal slice and an $8 \times 8$ spatial patch), which were projected into a 192-dimensional embedding space (for detailed explanations of deep learning related terms, see e.g., \citep{goodfellow2016:dl-book, prince2023:dl-book}). These embedded patches were then processed in two stages. In the first stage, a spatial encoder with two transformer layers and six-head self-attention was used to capture spatial features for each modality. In the second stage, a temporal encoder of identical design modeled temporal dependencies. The resulting spatio-temporal features from Sentinel-1 and Sentinel-2 were combined in a transformer-based decoder with cross-attention to encourage interaction between modalities. Finally, a multilayer perceptron (MLP) produced the segmentation outputs.


We trained the model using cross-entropy loss \citep{goodfellow2016:dl-book}, with each input channel normalized using robust statistics (centered by the median and scaled by the median-absolute-deviation (MAD)) computed from the training set. Pixels labeled as unknown were excluded from both loss and evaluation. Sentinel-1 and Sentinel-2 images were used jointly as inputs, and we applied CutMix augmentation \citep{yun2019cutmix}, in which a randomly generated rectangular region in one sample was replaced with the corresponding region from another, with the same mask applied to both the images and labels.
Model selection was based on the F1-score for tree crops on the test set. After determining the best configuration, we generated the final tree crop map using an ensemble approach. Specifically, we trained five instances of the proposed architecture, each initialized with a different random seed to improve robustness and reduce prediction variance. The logits from the five ensemble models were first averaged, and a softmax function was then applied to the averaged logits to obtain the final class scores (pseudo-probabilities). These scores were used to determine the final class prediction for each pixel by taking the class with the highest score, from which we derived the binary tree crop mask that constitutes the final tree crop map.


\subsection*{Accuracy assessment of the tree crop map}

To obtain unbiased estimates of class areas and rigorously quantify classification performance, particularly for rare classes susceptible to systematic bias, it is essential to employ design-based inference based on a probability sampling design \citep{olofsson2014good}. Because tree crops are a rare and spatially imbalanced class, direct use of mapped area estimates risks systematic bias. We therefore applied a probability-based accuracy assessment, following the sampling and estimation framework of \cite{olofsson2014good}.
We implemented a disproportionate stratified random sampling design, allocating an equal number of samples to tree crop and non–tree crop classes. The required sample size was calculated to achieve a target standard error ($SE = 0.005$) for the tree crop area estimate, assuming a user’s accuracy of 0.9 and a commission error of 0.1. This resulted in 3,510 reference samples, distributed across countries in proportion to their mapped tree crop area: 1,863 in Brazil, 601 in Colombia, 331 in Ecuador, 336 in Chile, 153 in Argentina, and 226 across other countries.
To further reduce bias from omission errors, which disproportionately affect rare classes \citep{olofsson2020mitigatingOmissionError}, an additional ``buffer'' stratum surrounding mapped tree crop areas was introduced, where such errors are most likely to occur \citep{arevalo2020rareClassREDD}. Buffer size was inversely related to national sample size (300 m for Brazil, 500 m for Colombia, 1,000 m for Ecuador and Chile, and 2,000 m for other countries). In each country, 100 samples were randomly allocated to this buffer stratum. This stratification effectively partitioned the large non–tree crop class into a smaller subset. %
All reference samples were manually labeled as ``tree crop'' or ``non-tree crop'' using very high-resolution satellite imagery from 2020 on Google Earth. Based on the resulting error matrix (Table \ref{tab:error_matrix}), we derived user’s and producer’s accuracies, overall accuracy. The map assessment results of all countries are summarized in Table \ref{tab:error_matrix}. From this error matrix, the overall accuracy was 99\%. For the tree crop class, the user’s accuracy (UA) was 82\% and the producer’s accuracy (PA) was 95\%.

The error-adjusted area estimates were further calculated only for countries where tree crops occupy more than 0.5\% of country area. This threshold, consistent with the imbalance level used by \cite{olofsson2020mitigatingOmissionError}, ensures stable and statistically reliable area estimates. In countries with lower tree crop coverage, extreme class imbalance introduces substantial uncertainty, and therefore these regions were excluded from the area adjustment analysis.
Following \cite{olofsson2014good}, the adjusted area $\hat{A}_i$ of class $i$ was estimated as:
\begin{equation}
    \hat{A}_i = A \cdot \sum_{h=1}^H W_h \hat{p}_{ih},
\end{equation}
where $W_h$ is the mapped area proportion of stratum $h$, and $\hat{p}_{ih}$ is the estimated proportion of reference class $i$ within stratum $h$, and $A$ is the total mapped area. The per-country estimates and their aggregated totals are presented in Extended Data Table~\ref{tab:area_estimate_country}.
 Across these countries, the total mapped tree crop area increased from 7.87 million ha to 9.91 $\pm$ 2.08 million ha after adjustment. We then derived a scaling factor ($\frac{\text{adjusted area}}{\text{initial area}} = 1.26$), which was applied to countries with stronger class imbalance to obtain consistent continental estimates. After applying this adjustment, the total estimated tree crop area across South America is 10.99 million ha.

\begin{table}[ht]
\caption{Error matrix for tree crop map assessment. Rows represent map predictions; columns represent reference labels. \(W_h\) is the stratum weight based on the mapped areas.}
\label{tab:error_matrix}
\begin{tabular}{@{}llccccc@{}}
\toprule
&  & \multicolumn{3}{c}{\textbf{Reference}} & \multicolumn{2}{c}{\textbf{Str.}}\\
\cmidrule(lr){3-5}\cmidrule(l){6-7}
 & \textbf{Stratum} &
\begin{tabular}[c]{@{}c@{}} Tree Crops \end{tabular} &
\begin{tabular}[c]{@{}c@{}} Non-Tree Crops \end{tabular}  & \textbf{Total} &
\begin{tabular}[c]{@{}c@{}}Str. area\\ {[}ha{]}\end{tabular} &
\begin{tabular}[c]{@{}c@{}}Str. weight\\ $W_h$\end{tabular} \\
\midrule
\multirow{3}{*}{\begin{tabular}[c]{@{}l@{}}M\\ a\\ p\end{tabular}}
 & \textit{Tree Crops} & \cellcolor{lightgreen}1277 & \cellcolor{lightblue}272 &  \textbf{1549} & 8,736,814 & 0.0047 \\
 & \textit{Non-Tree Crops}     & \cellcolor{lightred}4 & 1728 &  \textbf{1732} & 1,816,066,434 & 0.9801 \\
 & \textit{Buffer}         & \cellcolor{lightred}46 & 534  &\textbf{580} & 28,110,734 & 0.0152 \\
\midrule
 & \textbf{Total}          & \textbf{1327} & \textbf{2534} & \textbf{3861} & \textbf{1,852,913,982} & \textbf{1.000} \\
\bottomrule
\end{tabular}
\end{table}


It is important to note that this adjustment corrects area estimates but does not modify the individual pixel classifications of the map. Consequently, the total area calculated by summing the pixels in the map will differ from the unbiased, statistically robust area estimates reported here. Producing locally adjusted maps across South America would require country or region-specific reference datasets, especially in areas dominated by smallholder systems where locally collected information is often more reliable than remote sensing alone. Such refinements are beyond the scope of this study but represent an important direction for future research.

\section*{Data availability}

The tree crop map of South America is available in the Google Earth Engine App (\url{https://nature-trace.projects.earthengine.app/view/treecrops-v2}).
The validation datasets generated and analysed during the current study are available from the corresponding author on reasonable request.
The training dataset and the final map were generated using the \textit{GeeFlow} library  (\url{https://github.com/google-deepmind/geeflow}) that uses Google Earth Engine \cite{gorelick2017:gee} as the backbone.
The code for model training, inference, and evaluation is available in the \textit{JEO} code repository (\url{https://github.com/google-deepmind/jeo}).

\bibliographystyle{unsrt}
\bibliography{sn-bibliography}

\section*{Acknowledgments}
We thank Ștefan Istrate for support with data curation and Mélanie Rey for helpful review comments on the manuscript. We are also grateful to Louis Reymondin and Thibaud Vantalon from Alliance Bioversity International – CIAT for their review and assessment of the map products.
This project was partially funded by the Embed2Scale project, co-funded by the EU Horizon Europe Programme under Grant Agreement No. 101131841. Additional support was provided by the Swiss State Secretariat for Education, Research and Innovation (SERI) and UK Research and Innovation (UKRI).

\section*{Author Contributions}
Y.J. contributed to the conceptualization, methodology development, data analysis, validation, and visualization, and wrote the original draft with input from all co-authors. A.R. contributed to methodology, software development, and data processing. X.T. contributed to the initial draft and figure design. D.M. supported the conceptualization, data curation and manuscript editing. V.S.F.G., D.O.G., K.S., and J.D.W. provided critical feedback and revisions. M.N. led the conceptualization and methodology design, contributed to software development, investigation, and visualization, manuscript editing and supervised the project.

\ifArxiv
\else

\section*{Competing interests}
The authors declare no competing interests.

\fi

\section*{Extended Data}
\renewcommand{\thefigure}{E\arabic{figure}}
\setcounter{figure}{0}
\renewcommand{\thetable}{E\arabic{table}}
\setcounter{table}{0}

\begin{table}[H]
\centering
\caption{Tree crop area assessment by country. UA = User’s Accuracy and PA = Producer’s Accuracy (after adjustment). The initial area is derived directly from the map, while the adjusted area reflects post-adjustment estimates. $W_{tc}$ denotes the stratum weight of tree crops as a percentage of total land area.}
\label{tab:area_estimate_country}
\begin{tabular}{lcccccc}
\toprule
\textbf{Country} & \textbf{UA} & \textbf{PA} & \textbf{Initial area} & \textbf{Adjusted area} & \textbf{$W_{tc}$ (\%)} \\
\hline
Brazil & 90 & 62 & 4,609,541 & 6,644,558 & 0.528 \\
Colombia & 70 & 79 & 1,557,885 & 1,390,107 & 1.358 \\
Ecuador & 73 & 67 & 834,422   & 916,727   & 3.355 \\
Chile   & 76 & 68 & 871,309   & 967,329   & 0.922 \\
\midrule
\textbf{All countries ($W_{tc} > 0.5$)} & 83 & 66 & 7,873,157 & 9,908,008 & 0.711 \\
\bottomrule
\end{tabular}
\end{table}

\section*{Supplementary information}
\renewcommand{\thefigure}{A\arabic{figure}}
\setcounter{figure}{0}
\renewcommand{\thetable}{A\arabic{table}}
\setcounter{table}{0}

\subsection{Supplementary Figures and Tables}

\begin{figure}[!ht]
   \centering
        \includegraphics[width=.9\linewidth]{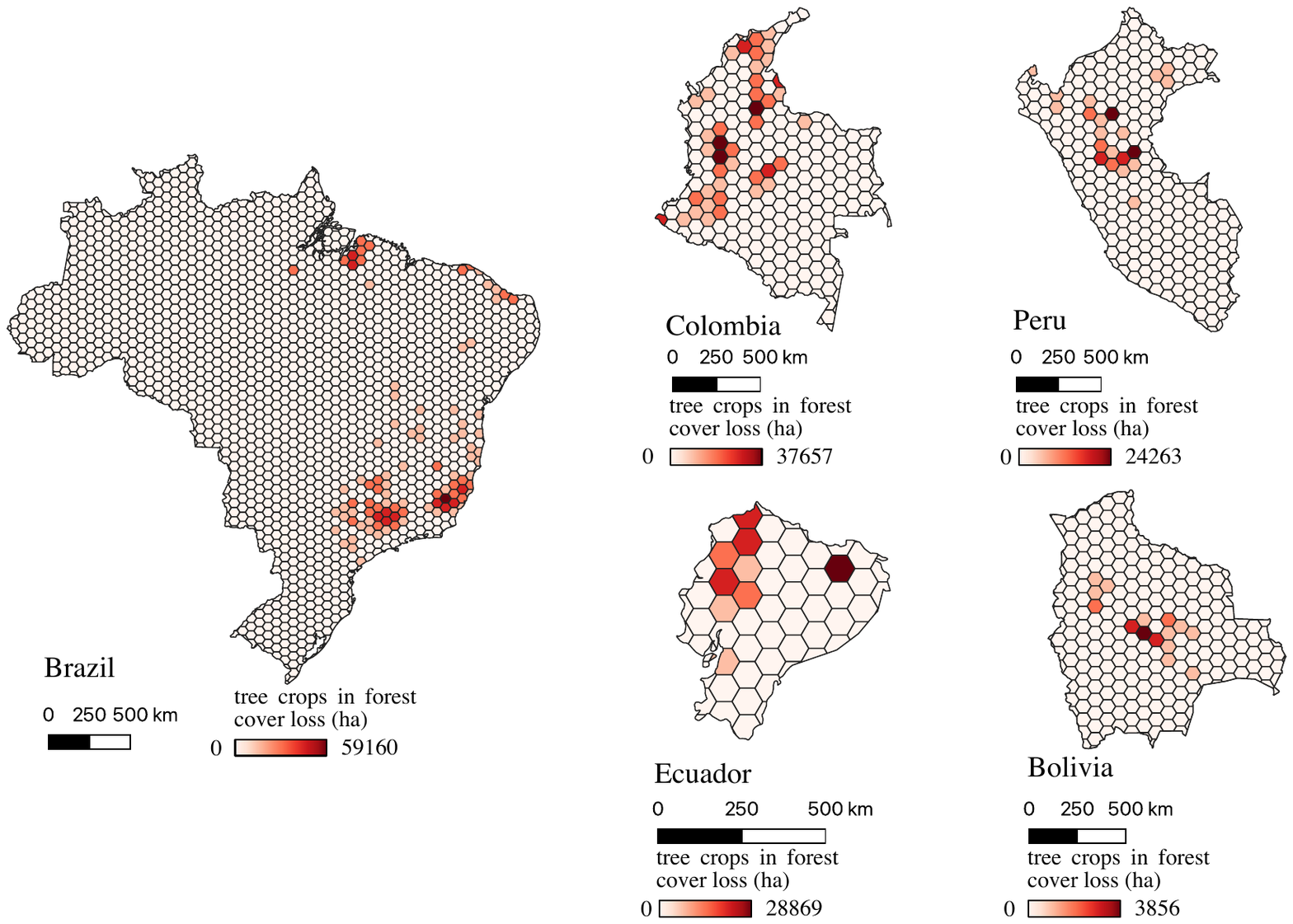}
        \caption{Hexagon map of tree crop areas inside forest cover loss per country. Each hexagon covers $80 \times 80$ $km^2$ area and darker hexagon indicate more tree crop areas inside forest cover loss. The line chart shows the annual tree crop area inside forest cover loss per country.}
        \label{supplementary_fig:forest_loss_spatial}
\end{figure}

\begin{itemize}
    \item Figure \ref{supplementary_fig:forest_loss_spatial}: Hexagon map of tree crop areas inside forest cover loss per country. 
    Each hexagon covers $80\times80~km^{2}$ area and darker hexagon indicate more tree crop areas inside forest cover loss.
    \item Table \ref{tab:tree-crop-sources}: Description of label sources used for the tree crop class. Also denoting the region (for regional dataset) and the number of sample locations.
    \item Table \ref{tab:other-sources}: Label sources for the non-tree crop classes.
    \item Figure \ref{supplementary_fig:sanity_check_tc_forest_loss}: Examples of visually interpreted tree crop areas within forest cover loss.
    \item Figure \ref{model_overview}: Overview of the model architecture.
    \item Figure \ref{fig:uncertainty_diagram}: Reliability diagram illustrating the relationship between predicted uncertainty and empirical accuracy.
    The purple line represents observed accuracy as a function of predicted uncertainty, while the dashed diagonal indicates perfect calibration. The Expected Calibration Error (ECE) is 8.1\%.
    \item Figure \ref{fig:confidence_overview_map}: Overview of the uncertainty map.
    A map showing the spatial distribution of prediction uncertainty (normalized entropy). Red areas indicate higher uncertainty (lower model confidence), while blue areas indicate low uncertainty. Uncertainty is highest in regions with complex landscapes, heterogeneous smallholder agriculture, and along the Andes.
\end{itemize}


\subsection{Visual interpretation check of tree crops within forest cover loss areas}\label{sanity_check_tc_forest_cover_loss}
\begin{figure}[!ht]
   \centering
        \includegraphics[width=.9\linewidth]{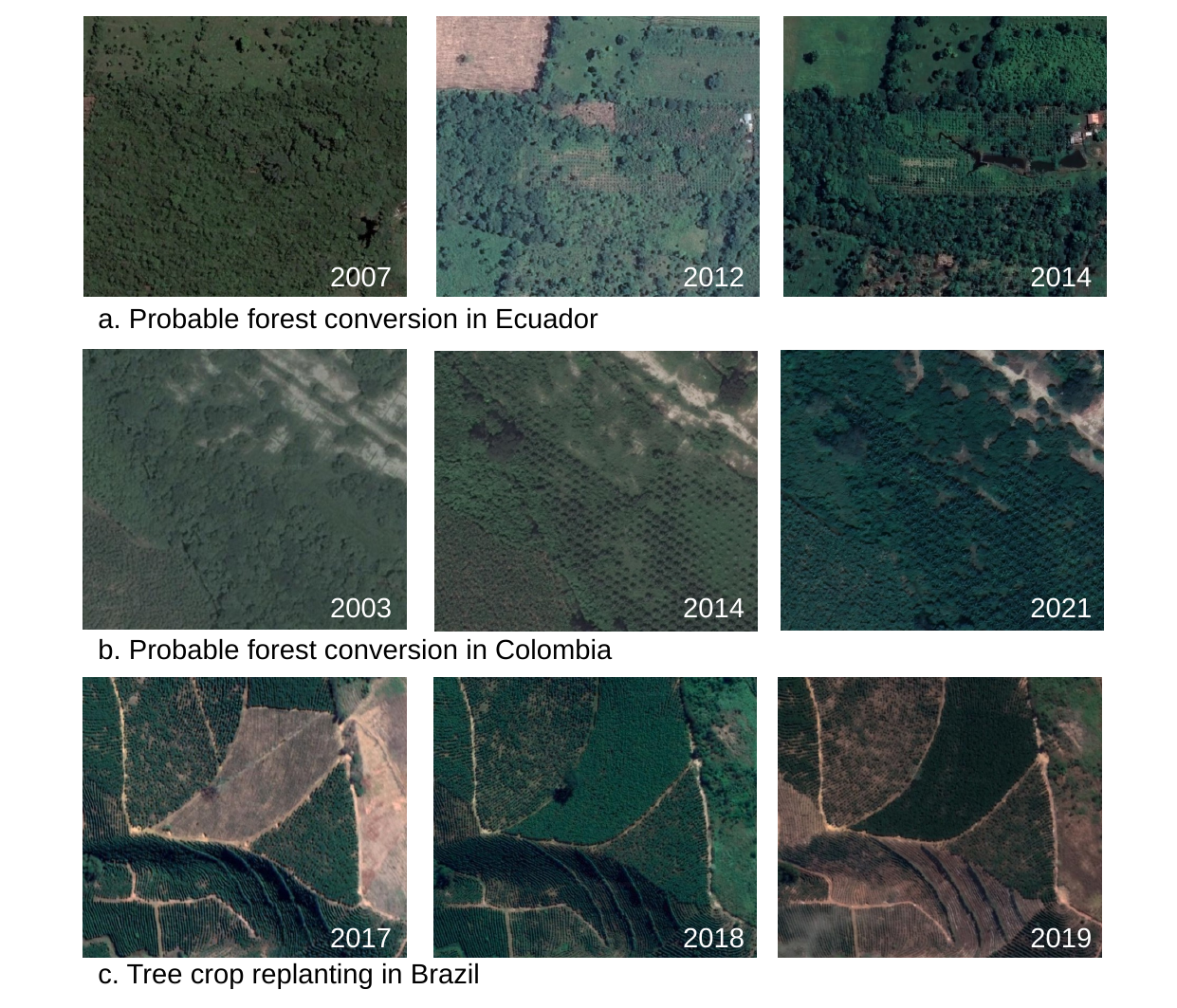}
        \caption{Examples of visually interpreted tree crop areas within forest cover loss. (a) and (b) show examples of likely forest-to-tree-crop conversion in Ecuador and Colombia, respectively, while (c) illustrates tree crop replanting in Brazil. Images are historical high-resolution satellite images from Google Earth.}
        \label{supplementary_fig:sanity_check_tc_forest_loss}
\end{figure}

To better interpret the overlap between tree crops and forest cover loss, we conducted a visual interpretation of 300 randomly sampled points from areas classified as tree crop within forest cover loss across Brazil, Colombia, Ecuador, and Peru, the four countries with the largest such overlaps. We focused on loss years after 2010, as earlier imagery in Google Earth was often unavailable or of insufficient quality for interpretation. 
For each sampled location, we examined historical high-resolution satellite imagery to determine whether the observed loss represented (1) probable conversion of forest to tree crop, (2) replanting of tree crops within previously cleared land, (3) potential false positives in the forest loss data \citep{hansen2013high_Hansen_forestloss}, where no canopy change was evident, (4) potential false positives in our tree crop map, where no tree crop was observed, or (5) uncertain or unclassifiable cases due to cloud cover or image quality.
Across all samples, 36\% corresponded to clear forest-to-tree-crop conversion, 35\% to replanting, 5\% to potential false positives in the forest loss dataset, 1\% to potential false positives in our tree crop map, and 23\% remained uncertain. Examples of visual interpretation are shown in Figure \ref{supplementary_fig:sanity_check_tc_forest_loss}. It is noticeable that in the tree crop replanting cases are mostly smallholder plantations instead of industrial-scale plantations like large-scale oil palm plantations. Large-scale plantations like oil palm, tend to be stable, with forest loss corresponding mainly to new establishment rather than rotation.
These findings highlight that canopy loss identified in satellite-based forest loss products does not always represent new deforestation; in some cases, replanting or management cycles of tree crops can also appear as tree cover removal in such datasets \citep{hansen2013high_Hansen_forestloss}.

\subsection{Datasets}

\begin{table}[!ht]
    \centering
    \caption{Description of label sources used for the tree crop class. Also denoting the region (for regional dataset) and the number of sample locations.}
    \label{tab:tree-crop-sources}
    \begin{tabular}{p{1.2in}|p{4in}}
    \toprule
    \textbf{Name} & \textbf{Description} \\
    \midrule
    SDPT Tree Crops & A collection of tree crop sources, including both, manual annotations and inference results. (1,222 sample locations collected during 2000 - 2020 )\citep{richter2024SDPTv2}.\\
    GFM & Tree crop (palm, rubber, orchards) training samples (3,557 sample locations in 2015) \citep{lesiv2022global_lesiv2015}.\\
    SERVIR Ucayali Palm & Annotated palm polygons in Peru (205 sample locations in 2015) \citep{fricker2022palmUcayali}.\\
    Palm Descals & Annotated palm polygons (66 sample locations in 2019) \citep{descals2024global_palmDescals}.\\
    MapBiomas & A collection of tree crop sources (can include manual annotations and inference results) in Brazil (14,971 sample locations collected during 2016 to 2019) \citep{souza2020reconstructingMapBiomasBrazil,mapbiomas8}.\\
    Palm Vollrath & Oil palm annotations (104 sample locations in 2017) \citep{vollrath2020angular_palmVollrath}.\\
    Coconut Descals & Coconut palm annotations (1 location in 2020) \citep{descals2023high_coconutDescals}.\\
    SERVIR Cocoa & Annotated cocoa polygons in Peru (87 sample locations in 2020) \citep{becerra2022:servir-cocoa-peru}.\\
    Manually annotated data & Manually annotated (by experts or crowd-sourcing) (2,702 sample locations in 2020).\\
    \bottomrule
    \end{tabular}
\end{table}

\begin{table}[!ht]
    \centering
    \caption{Label sources for the non-tree crop classes.}
    \label{tab:other-sources}
    \begin{tabular}{p{1.2in}|p{4in}}
    \toprule
    \textbf{Name} & \textbf{Description} \\
    \midrule
    SDPT Planted forest & A collection of planted forest sources, including both, manual annotations and inference results \citep{richter2024SDPTv2}.\\
    GFM & Naturally regenerating and planted/managed forest training samples\citep{lesiv2022global_lesiv2015}.\\
    TMF & Tropical moist forest inference results \citep{vancutsem2021long_TMF}.\\
    SBTN & A collection of natural and non-natural land \citep{Mazur_Sims_Goldman_Schneider_Pirri_Beatty_Stolle_Stevenson_SBTN}.\\
    WorldCover & Inference map of land cover land use classes \citep{zanaga2022esaWorldCover}.\\
    Primary Humid Tropical Forests & Inference map of natural forests \citep{turubanova2018PrimaryForest}.\\
    MapBiomas & Inference map of land cover \citep{souza2020reconstructingMapBiomasBrazil} to map savanna lands.\\
    LCZ & Local climate zone map \citep{demuzere2022globalLCZ} to disambiguate bushes.\\
    \bottomrule
    \end{tabular}
\end{table}

\subsection{Deep learning framework}
\begin{figure}
  \centering
  \includegraphics[width=\textwidth]{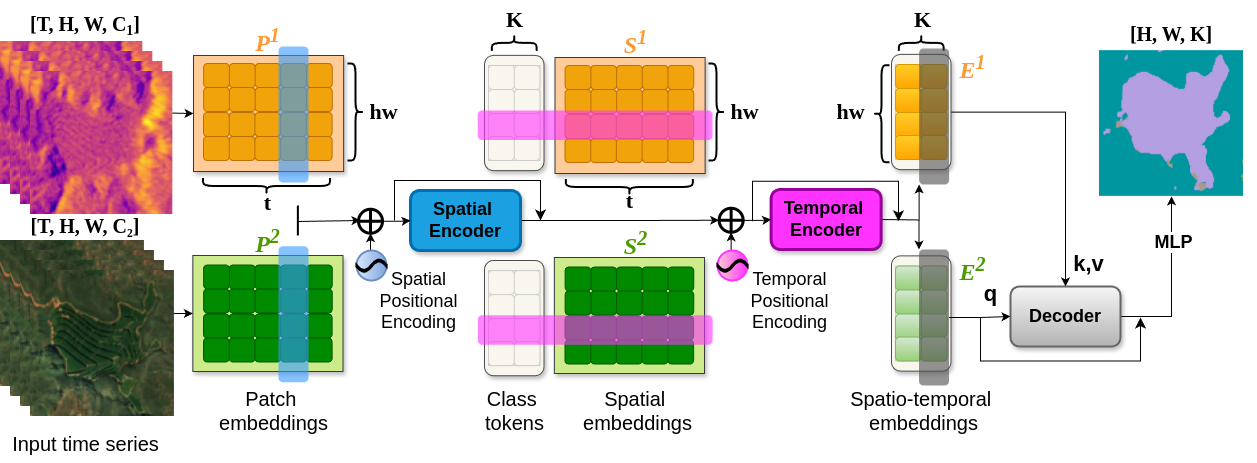}
  \caption{\textbf{Overview of the proposed method.}
  The input consists of tensors from Sentinel-1 ($[T, H, W, C_1]$) and Sentinel-2  ($[T, H, W, C_2]$). The model generates a segmentation map of shape  ($[H, W, K]$, $K$ is the number of classes). The dimensions handled by spatial encoder, temporal encoder and the decoder are highlighted in blue, purple, and gray, separately.
  }
  \label{model_overview}
\end{figure}

The detailed model architecture is in Figure \ref{model_overview}. Each input image time series obtained from Sentinel-1 and Sentinel-2, and with respective shape $T \times H \times W \times C_m$, with $T$ the temporal sequence length, $H\times W$ the spatial extent in pixels, and $C_m$, the modality-specific number of channels. 
Our spatial and temporal encoders followed the transformer design of \cite{vaswani2017attention}, each consisting of two layers with six attention heads and an embedding size of 192. The multi-modal decoder was a two-layer transformer decoder with the same embedding size, and residual connections were applied throughout. In total, the model comprised 3.4 million parameters, making it relatively lightweight.

Training of this model was implemented in JAX \citep{jax2018github}, using a batch size of 64 and an initial learning rate of 0.001. Optimization was performed with Adam \citep{kingma2014adam} ($\beta_1$ = 0.9, $\beta_2$ = 0.999, $\epsilon = 10^{-8}$), and a cosine annealing schedule \citep{loshchilov2016cosineAnneal} was used to adjust the learning rate over 40 epochs. 
\subsection{Uncertainty quantification}

Quantifying prediction uncertainty is essential for interpreting large-scale land cover maps, as it provides users with a measure of confidence in individual predictions and highlights regions where caution is warranted. We quantified pixel-level uncertainty using entropy, a standard measure derived from the model’s class probabilities:
\begin{equation}
    H = -\sum_{i=1}^{K} p_i \log p_i
\end{equation}
where \( H \) represents the entropy, \( K \) is the number of classes, and \( p_i \) is the predicted probability of class \( i \). For interpretability, entropy was normalized to the range $[0,1]$ and used as our uncertainty measure. 
To assess how well this uncertainty aligns with actual model performance, we evaluated calibration using the Expected Calibration Error (ECE) based on the independent reference labels created for accuracy assessment. ECE measures the gap between predicted confidence and observed accuracy across a set of bins:
\begin{equation}
\text{ECE} = \sum_{m=1}^{M} \frac{|B_m|}{n} \left| \text{acc}(B_m) - \text{conf}(B_m) \right|
\end{equation}
where $M$ is the number of uncertainty bins (we choose 12 bins), $B_m$ is the set of samples in bin $m$, $|B_m|$ is the number of samples in that bin, $n$ is the total number of samples. $\text{acc}(B_m)$ was computed by comparing map predictions to reference labels, while $\text{conf}(B_m)$ was the average model confidence ($1 - normalized\ entropy$) for that bin. A lower ECE indicates better alignment between predicted uncertainty and actual accuracy, and thus better calibration of the uncertainty estimates.

The resulting ECE of 8.1\% indicates good calibration quality, showing that entropy-based uncertainty corresponds well with empirical accuracy. This supports the use of prediction entropy as a meaningful quality flag for the final tree crop map, enabling users to identify regions where predictions are less reliable. The reliability map and the overview of the uncertainty map of South America are provided in the Supplementary Figure \ref{fig:uncertainty_diagram} and \ref{fig:confidence_overview_map}.

\begin{figure}[ht]
   \centering
        \includegraphics[width=.5\linewidth]{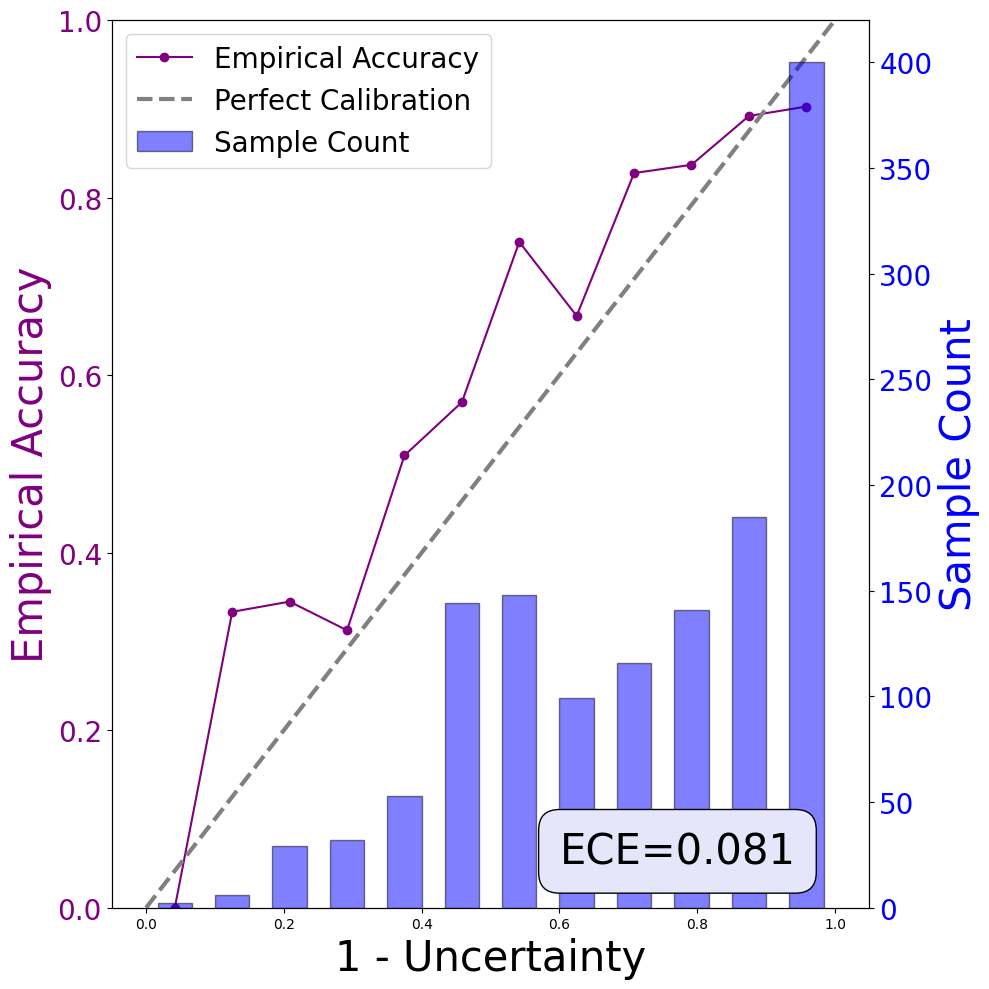}
        \caption{Reliability diagram illustrating the relationship between predicted uncertainty and empirical accuracy. The purple line represents observed accuracy as a function of predicted uncertainty, while the dashed diagonal indicates perfect calibration. The Expected Calibration Error (ECE) is 8.1\%.}
        \label{fig:uncertainty_diagram}
\end{figure}

\begin{figure}[htp]
   \centering
        \includegraphics[width=\linewidth]{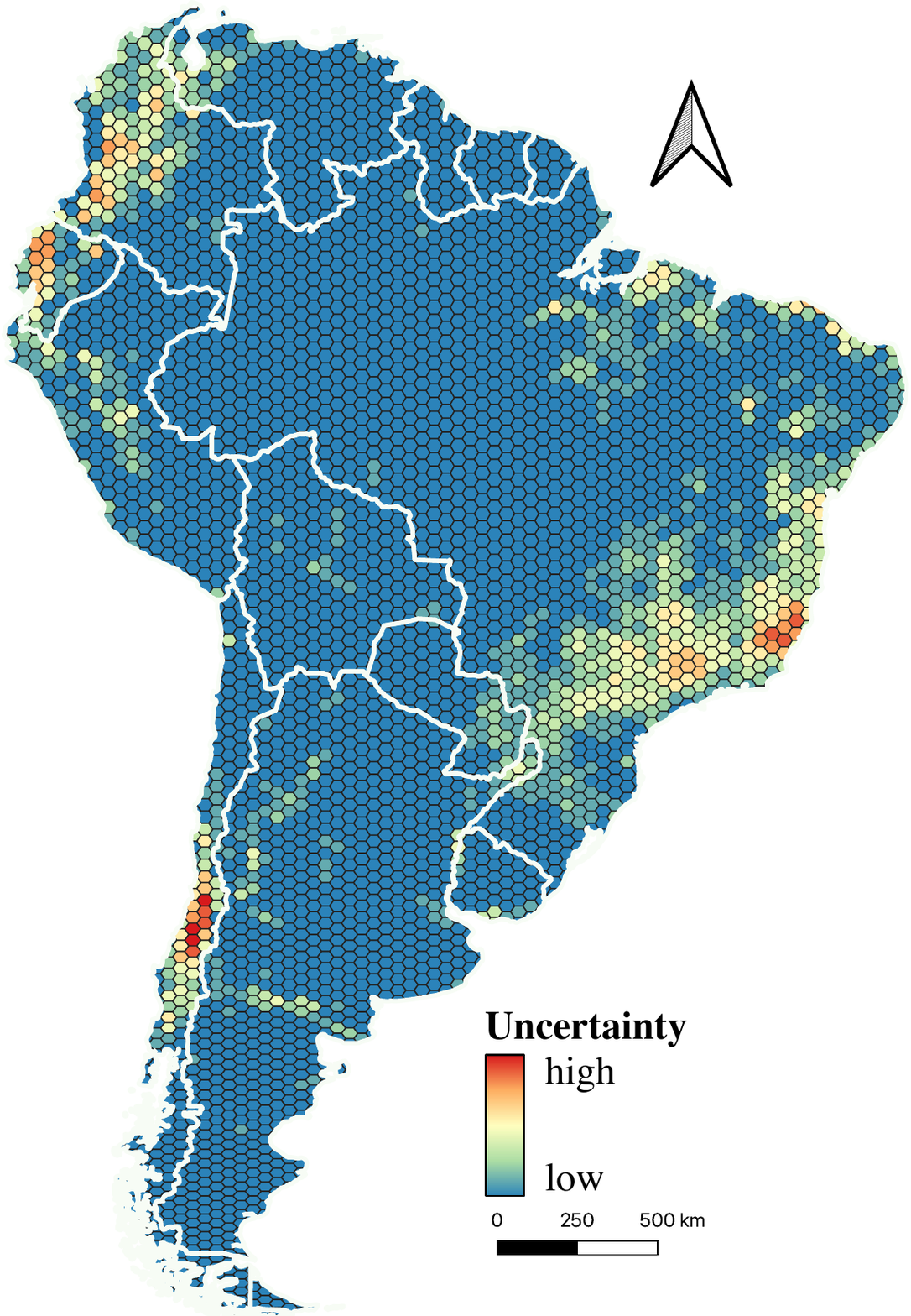}
        \caption{Overview of the uncertainty map}
        \label{fig:confidence_overview_map}
\end{figure}

\end{document}